\documentclass[conference]{IEEEtran}
\usepackage{times}

\usepackage[numbers]{natbib}
\usepackage{multicol}
\usepackage[bookmarks=true]{hyperref}
\usepackage{graphicx}

\begin{document}

% paper title
\title{Real-Time Escape Route Generation in Low Visibility Environments using Reinforcement Learning}

% You will get a Paper-ID when submitting a pdf file to the conference system
\author{Author Names Omitted for Anonymous Review. Paper-ID [add your ID here]}

\author{\authorblockN{Hari Srikanth}
\authorblockA{Mira Loma High School \\
Sacramento, California 95761\\
Telephone: (916) 693-0389 \\
Email: harinsrikanth@gmail.com}
}

\maketitle

\begin{abstract}
Structure fires are responsible for the majority of fire-related deaths nationwide. In order to assist with the rapid evacuation of trapped people, this paper proposes the use of a system that determines optimal search paths for firefighters and exit paths for civilians in real time based on environmental measurements. Through the use of a LiDAR mapping system evaluated and verified by a trust range derived from sonar and smoke concentration data, a proposed solution to low visibility mapping is tested. These independent point clouds are then used to create distinct maps, which are merged through the use of a RANSAC based alignment methodology and simplified into a visibility graph. Temperature and humidity data are then used to label each node with a danger score, creating an environment tensor. After demonstrating how a Linear Function Approximation based Natural Policy Gradient RL methodology outperforms more complex competitors with respect to robustness and speed, this paper outlines two systems (savior and refugee) that process the environment tensor to create safe rescue and escape routes, respectively.

\end{abstract}

\IEEEpeerreviewmaketitle

\section{Motivation}
Structure fires pose a serious danger to many communities across the world, and this problem has only grown over the past decade, with over a 33\% increase in deaths from fire since 2012 (\citet{fire}). Of these, structure fires are most prominent, constituting 79\% of civilian deaths and 86\% of civilian injuries. These casualties stem from people being trapped within the building for an extended period of time, resulting in exposure to collapsing structures, high heat/flames, and toxic smoke. In addition, the unpredictable nature of fires reduces the efficacy of predetermined escape strategies. In order to determine safe rescue and escape routes, a live mapping of the environment with the measurement of aspects of interest is necessary. However, current methodologies are insufficient to handle this task, due to an inability to map rapidly in densely obstructed environments, as well as slow onboard machine learning. 

\section{Background}
The first task for the system is to generate a map of the environment in real time. A well researched method for this is Simultaneous Localization and Mapping (SLAM). SLAM is a process where the agent pose is determined within an environment which is (simultaneously) being mapped. While SLAM methodologies have been well tested in regular environments, low visibility environments clouded with smoke or dust pose a problem for most systems, as they utilize optical rangefinding systems. In order to develop a resilient mapping system, various solutions have been proposed:
\begin{itemize}
    \item SLAM with Visual and Thermal Imaging Cameras (\citet{BrunnerUGV}): Utilizes thermal imaging in order to counterbalance visual camera obfuscation. However, increased reliance on thermal data resulted in a decrease in localization accuracy.
    \item SLAM with laser range finders and 94 GHz Frequency Modulated Continuous Wave Radar (\citet{Castro}): Proposed a system with mm-wave radar to penetrate dense smoke that laser scanners cannot. Very precise, but some false negatives remained. In addition, the custom antenna system is very costly and locks mobility.
    \item SLAM with laser range finders and sonar sensor array (\citet{SantosSonarLRF}): Uses a fuzzy logic system \citet{Couceiro20121625} to decide between sensor inputs, reducing the risk of false negatives.
\end{itemize}
Regardless of the specific mapping methodology, the map resolution and effective range drastically decrease as the adversarial noise parameter increases. Given these limitations, the use of Multi-Robot Systems (MRS) is necessary to optimize mapping speed and robustness, with a Random Sample Consensus (RANSAC) based map unification to ensure live processing does not become too computationally expensive (\citet{LazaroMLS}). 
The second aspect of the problem is the determination of optimal rescue or escape paths. In order to do this, the situation can be modeled through a Markov Decision Process and solved using Reinforcement Learning. Reinforcement Learning (RL) is a paradigm where an agent seeks to maximize the reward it gains through refining its policy. At each timestep t, the agent observes the environmental state and according to some policy $\pi$ it takes some action. This action changes the environmental state and returns some reward, and this is used to retrain the policy. A well researched RL method is Actor Critic, which utilizes an actor that updates the policy and a critic which evaluates said policy for fast optimization. While industry standard algorithms Trust Region Policy Optimization (\citet{TRPO}) and Proximal Policy Optimization \citet{PPO} utilize complex neural networks to estimate the value function, my prior research on reinforcement learning determined that a Linear Function Approximation based Natural Policy Gradient algorithm (LFA-NPG) would determine the optimal policy with much less iterations than either TRPO and PPO, in low dimensional standard and sparse reward RL benchmarks. In addition, my robustness analysis determined that LFA-NPG was significantly more noise resistant to adversarial noise than TRPO and PPO, maintaining identical performance across data sampled within 20\% of the true value. Although the determination of navigation paths appears to be optimized for operations research methodology, the unique nature of each fire does not lend itself to the determination of the heuristics often employed in these algorithms. 

\section{Design}
The main necessities of a real world robotics system are as follows: 
\begin{itemize}
    \item Speed: Data must be sampled at a high rate, in order for the system to react appropriately
    \item Robustness: Algorithms must be resistant to adversarial noise, an inherent factor when live systems are considered
\end{itemize}
In addition to these two criteria, the search and rescue application motivating this paper requires the system to have some measure of its own accuracy. Fig. 1 details the proposed system, taking these criteria into consideration.

\begin{figure}[h]
    \centering
    \includegraphics[width=7 cm]{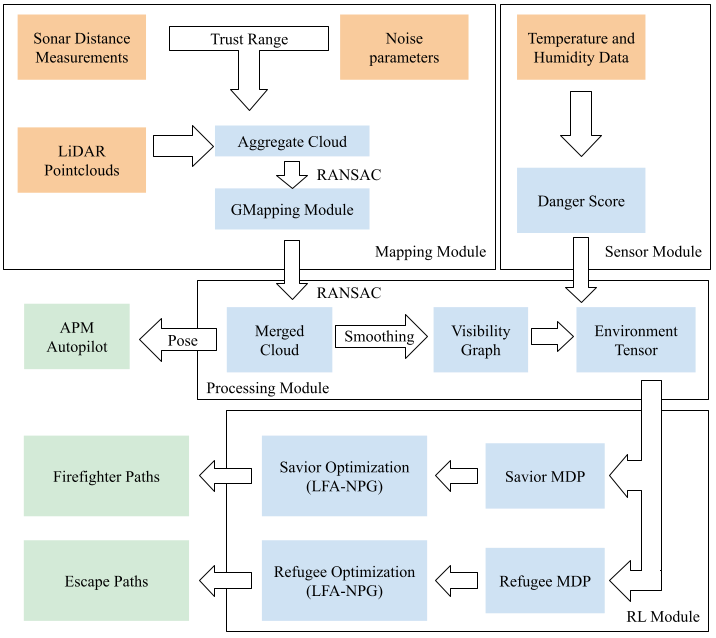}
    \caption{System Organization}
    \label{fig:enter-label}
\end{figure}

\subsection{Data Acquisition \& Processing}
Given that each second the system takes to find an exit path is a second of exposure to hazardous conditions, the key priority of my solution is speed. Fast navigation, fast mapping, fast processing, and fast path determination take precedence over exact optimization or high resolution. In order to map at this rate, the system will utilize a fleet of autonomous drones. Each drone will be outfitted with a mapping module (consisting of a LiDAR rangefinder and four sonar scanners) and a sensor module (to collect temperature and humidity data, as well as determine a quantitative adversarial noise parameter). Using the temperature and humidity data, the map will then be populated with scores quantifying the danger of each area of the building. From here, the LFA-NPG agent will determine the optimal paths, both for rescue (for the firefighters) and for escape (for the inhabitants). 
The base hardware for each device is a light quadrotor with ROS on a RPi running ubuntu. Using 3D printed parts to keep the weight as low as possible, the drone will have a target thrust/weight of 2:1. Each RPi interfaces with a Navio 2 flight control unit running an ardupilot APM autopilot. ROS features a variety of nodes to run SLAM calculations, such as HectorSLAM, GMapping, KartoSLAM, CoreSLAM, and LagoSLAM. Of these, GMapping is the most optimal for mapping in a smoky environment, as it remains robust and has a large amount of support.

The raw data that the mapping module collects from each drone is laser scan information from the LiDAR, four sonar distance measurements, and a measure of adversarial noise. The core of the mapping system is the LiDAR data: even though its accuracy is compromised in smoky environments, it is the most high-resolution sensor available within the price range. The sonar data comes from four sensors at 90 degrees to one another. The sonar data is then transformed to a laser scan frame to directly evaluate it in comparison to the LiDAR data. In order to determine which sensor data should be used (LiDAR for high resolution vs Sonar for higher accuracy) when constructing the map, the device incorporates a fuzzy logic system, through a measurement of confidence called the Trust Range. The trust range is determined by the difference between laser and sonar data and the adversarial noise parameter. In a design intended for real world use, the adversarial noise parameter in the use case of a fire would be determined by a particulate matter sensor. However, the logistics of producing real smoke multiple times for testing is a safety concern, and as such the effect of smoke is simulated through the use of a smoke machine. Because smoke machines do not produce particulate smoke, but rather vaporize alcohol-based fog solution, the noise parameter is instead determined through the use of an alcohol sensor (no difference in algorithm, just uses a different function and hardware when calculating). After the trust range is determined, the points outside the trust range are eliminated from the data and points within are added to an aggregate point cloud. This aggregate point cloud includes data from each drone exploring the system. The data from each point cloud is then passed into GMapping, creating a map. The set of maps from each drone is then fed into the processing module.

The processing module begins by merging each of the individual maps together, using the algorithm outlined below:
\begin{enumerate}
    \item Use a correlative scan matcher to identify matching edges of each scan
    \item  Add each edge solution to a pool of candidates
    \item For any two maps, determine a translation between them such that one candidate edge is identical (zero error).
    \item Determine the errors of all other candidate edges. Depending on if the error of the other edges is high or low, it is possible to determine inliers and outliers
    \item Once the correct translation has been determined, merge the two maps into a global map. This global map can then be merged with the next map, which can repeat until all maps have been integrated into the global map.
\end{enumerate}
After the global map has been created, the pose of each drone is passed to the flight controller, completing the localization aspect of SLAM. While this map could just be exported as is, more processing and simplification is necessary to determine exit paths. First, the merged map is smoothed into a visibility graph, which maps complex environmental features to a significantly more condensed set of traversable nodes. Each node is then assigned a danger score based on temperature and humidity data. For nodes where no temperature data has been collected, data from nearby nodes is weighted and used to estimate the conditions at the location. The complete environment tensor with traversable nodes and dangers is then ready to be fed into the RL training models.
\subsection{RL Data Analysis}
LFA-NPG is an policy-space RL Paradigm that utilizes the Natural Actor Critic (\citet{PETERS20081180}) Architecture, optimizing the policy through natural gradient descent while estimating the value function (how well a given policy will produce reward) through Linear Function Approximation. LFA-NPG was evaluated on standard RL benchmarks, Cartpole, and Acrobot (a sparse-reward environment, which is more common in real world robotics systems). Its performance is as shown in Fig. 2.

\begin{figure}[h]
    \centering
    \includegraphics[width=7 cm]{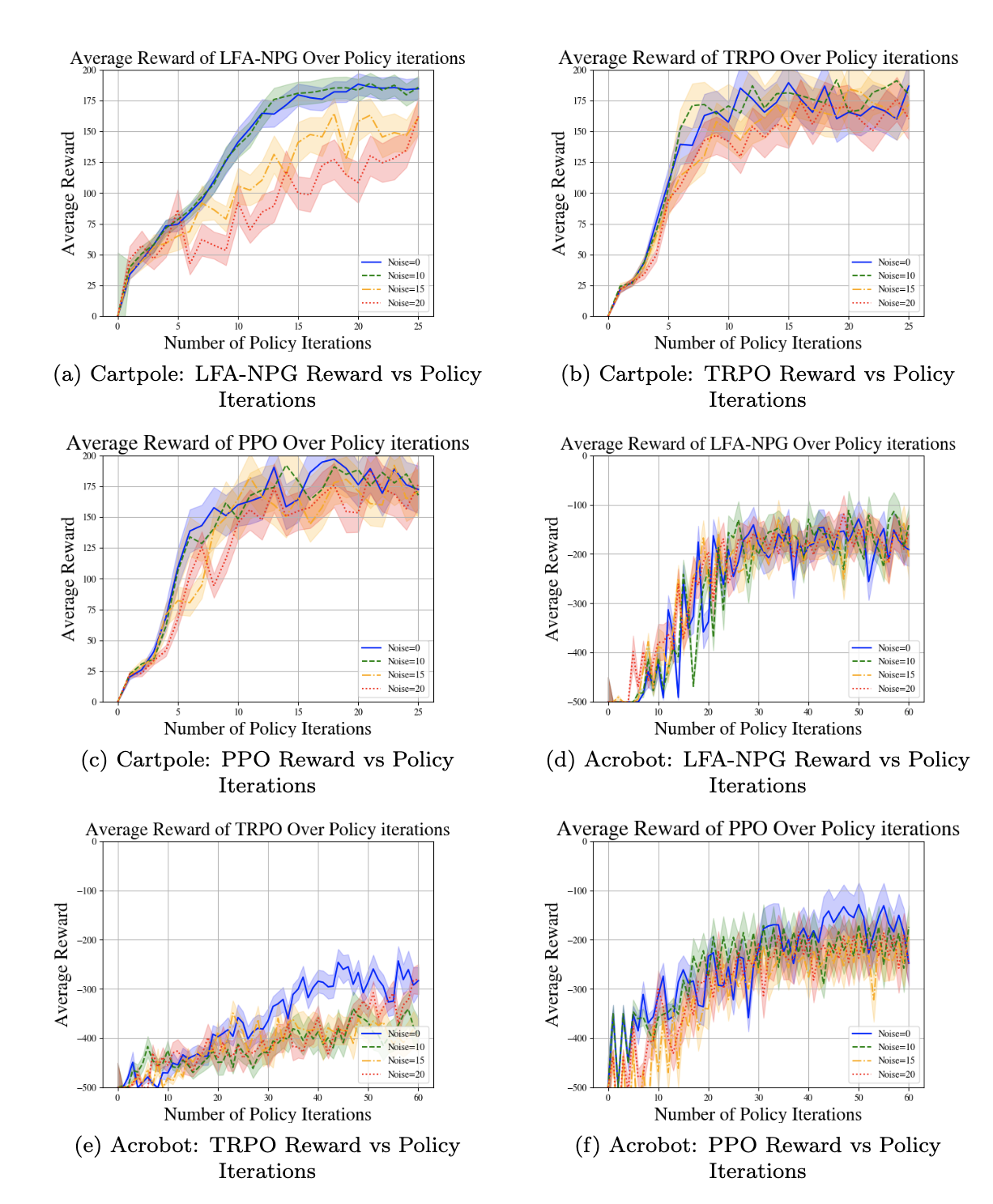}
    \caption{LFA-NPG Robustness Analysis}
    \label{fig:enter-label}
\end{figure}

As shown, LFA-NPG converges to the optimal policy even as the noise parameter (represented by the true state sampling error passed through to the model) increases up to 20\%. This is due to its simple value estimation method: in contrast to the complex neural networks present in other forms of RL, LFA-NPG's simpler value estimation methodology enables it to be more resistant to fluctuations in input.
In addition to robustness, LFA-NPG also is much faster than competing algorithms in low dimensionality applications (such as CartPole and Acrobot). It reaches a similar level of reward compared to PPO and TRPO, with a much faster and more logarithmic convergence, ideal for high speed use cases. This is evidenced in Fig. 3.

\begin{figure}[h]
    \centering
    \includegraphics[width=7 cm]{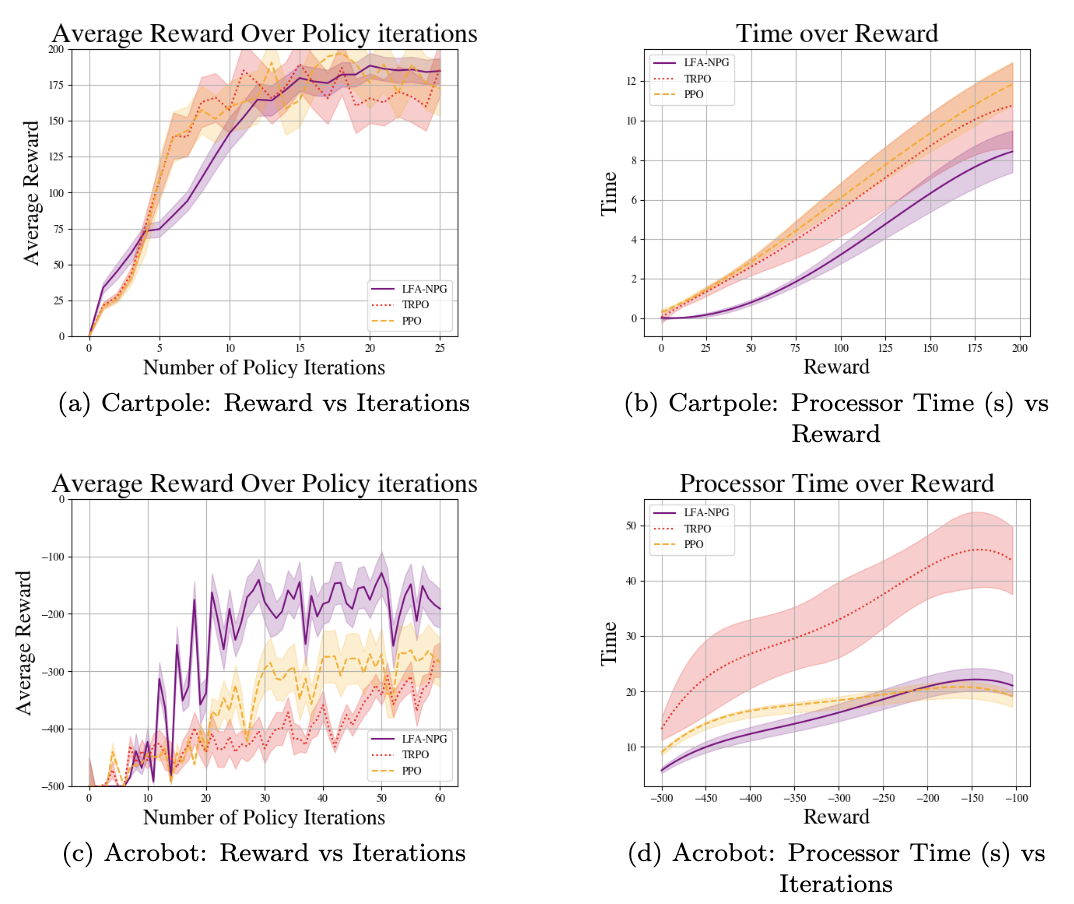}
    \caption{LFA-NPG Model Convergence w.r.t Iterations \& Time}
    \label{fig:enter-label}
\end{figure}

Considering that the data that can be collected during a live system activity is limited in scope, it will tend to be sufficiently low dimensionality enough for LFA-NPG to function at an optimal level. Moreover, RL methodology has been shown to be competitive with leading operations research algorithms currently used for traversal problems, and since the structure of an MDP is more cohesive with raw sensor inputs than the heuristic information necessary for OR algorithms, it is uniquely positioned as the optimal real time path planner for real world robotics. 

In the presented use case, the environment tensor can be used to inform both the optimal strategy for a firefighter and the optimal strategy for a trapped civilian. As such, two RL models will train simultaneously using the same base state space, with different reward functions. The savior system determines the strategy for the firefighter: Given the locations of possible entryways into the building, the system then determines the optimal paths to navigate to any point within the building, in the form of a tree. The refugee system is used to determine the escape routes. The starting nodes are all lethal areas (determined by danger score) and have a reward function dependent on duration and level of exposure to hazardous conditions. These two objectives are each used to formulate distinct MDPs, which LFA-NPG based solvers optimize through the use of the actor critic architecture. 

In order to determine the feasibility of this technology in real world applications, a minimum of 2 drones must be utilized (to evaluate the efficacy of the merged map) in a simulated indoor fire. This indoor fire will be simulated through the use of a smoke machine, which the glycol sensor uses to determine the danger parameter. As it is difficult to simulate the high temperatures of a fire, the environment tensor will instead be fed a possible matrix of danger scores within the building.

\section{Conclusion} 
\label{sec:conclusion}
This work considers the relevant Search and Rescue use case for autonomous robot systems for the traversal of a burning building. In addition to discussing the efficacy of current low visibility perception systems, this work proposes a novel perception system that makes use of a multi agent mapping array. Each individual agent prioritizes certain flows of data through a trust range, determined by the measure of adversarial noise, which is additionally used to evaluate the safety of each location in the map. It goes on to discuss a possible way in which this data can be analyzed, namely for the generation of escape/rescue routes that maximize safety. In order to accomplish this, the paper recommends the use of a linear function approximation based natural policy gradient reinforcement learning methodology, demonstrating its high speed and strong resistance to adversarial noise in sufficiently low dimensional systems.

%% Use plainnat to work nicely with natbib. 

\bibliographystyle{plainnat}
\bibliography{references}

\end{document}